\documentclass[letterpaper, 10 pt, conference]{ieeeconf}

\IEEEoverridecommandlockouts

\usepackage{amsmath,amsfonts,amssymb}
\usepackage{graphicx}
\usepackage{makecell}
\usepackage{mathtools}
\usepackage{subcaption}
\usepackage[table]{xcolor}
\usepackage{todonotes}

\usepackage{booktabs}
\usepackage{multicol}
\usepackage{multirow}
\usepackage{makecell}
\usepackage{siunitx}

\makeatletter
\newcommand{\vast}{\bBigg@{4}}
\newcommand{\Vast}{\bBigg@{5}}
\makeatother

\DeclareCaptionLabelSeparator{periodspace}{.\quad}
\title{\LARGE \bf
Learning Whole-Body Human-Robot Haptic Interaction\\ in Social Contexts
}

\author{Joseph Campbell$^{1}$ and Katsu Yamane$^{2}$%
\thanks{$^{1}$School of Computing, Informatics, and Decision Systems Engineering, Arizona State University. This work was done while J. Campbell was with Honda Research Institute, USA.}%
\thanks{$^{2}$Honda Research Institute, USA}%
}

\begin{document}

\maketitle
\thispagestyle{empty}
\pagestyle{empty}

\begin{abstract}
This paper presents a learning-from-demonstration (LfD) framework for teaching human-robot social interactions that involve whole-body haptic interaction, i.e. direct human-robot contact over the full robot body.
The performance of existing LfD frameworks suffers in such interactions due to the high dimensionality and spatiotemporal sparsity of the demonstration data.
We show that by leveraging this sparsity, we can reduce the data dimensionality without incurring a significant accuracy penalty, and introduce three strategies for doing so.
By combining these techniques with an LfD framework for learning multimodal human-robot interactions, we can model the spatiotemporal relationship between the tactile and kinesthetic information during whole-body haptic interactions.
Using a teleoperated bimanual robot equipped with 61 force sensors, we experimentally demonstrate that a model trained with 121 sample hugs from 4 participants generalizes well to unseen inputs and human partners.
\end{abstract}

\section{Introduction}

Physical human-robot interaction (pHRI) in social contexts requires the robot to be cognizant of the safety and comfort of the human partner.
Whole-body haptic information, consisting of both kinesthetic and tactile information, is critical in intimate haptic interactions as the arms are often occluded and touch is the only valid source of information on the timing and intensity of contact. 
In hugging, for example, each person feels the other's hugging force at the chest and back and adjusts their own exerted force in order to reciprocate.

While whole-body haptic modalities are starting to see usage in pHRI scenarios, they have thus far been largely limited to state classification and behavior selection using a snapshot of tactile information~\cite{argall2010survey}.
The main difficulties in using continuous-time tactile information for learning haptic interactions are twofold: 1) high data dimensionality and 2) temporal and spatial sparsity.
Whole-body tactile sensing requires tens or even hundreds of sensors to cover a wide area while preventing gaps in coverage.
The high dimensionality of tactile data not only increases the computational complexity but also makes the effect of tactile information on the model larger than that of kinesthetic information.
Another negative effect of dense sensor placement is that the data of individual sensors tends to be both temporally and spatially sparse because tactile information is relevant only when the robot and human partner are in contact, and many of the sensors may not see any contact during a specific interaction.
These difficulties call for a feature selection technique to efficiently handle high-dimensional, sparse tactile information.

In this paper, we develop three techniques for reducing the dimensionality of tactile information for learning haptic pHRI through a learning-from-demonstration (LfD) framework.
Firstly, we take advantage of temporal sparsity by employing a non-uniform distribution of basis functions to reduce the dimensionality of the latent space that represents the demonstrated interactions.
The second technique leverages spatial sparsity to choose the relevant tactile sensors by maximizing the mutual information between the input and output force measurements, effectively choosing the sensors which exhibit the highest mutual dependency.
An issue with this method is that due to variations in human motion and body size, the chosen sensors may not experience contact during actual interactions.
To remedy this, our third technique uses hand-crafted features for inference; specifically, the maximum force value within a predefined (local) tactile sensor group.
This method is motivated by the intuition that there will be no significant difference in perception regardless of where contact occurs in a densely packed sensor group, i.e. the force corresponding to a general spatial location is more important than a specific one.

We model the interaction as a Bayesian Interaction Primitive (BIP)~\cite{campbell2017bayesian, campbell2019probabilistic, campbell2019learning}, a spatiotemporal LfD framework.
This model is capable of predicting both an appropriate robotic response (consisting of joint trajectories) as well as contact forces that should be exerted by the robot, given observations of the partner's pose and the force currently exerted by the partner onto the robot.
The predicted joint trajectories and contact forces are then sent to a motion retargeting controller~\cite{kaplish2019motion}, which is capable of accounting for variances in the partner's body shape and size.
While previous applications of this framework include haptic interaction~\cite{campbell2019probabilistic}, the tactile information has thus far been dense, low-dimensional, and only used as an input observation.

The contributions of this work are summarized as follows:
\begin{itemize}
    \item a pHRI framework that combines the prediction of robot joint trajectories and contact forces with a motion retargeting controller to realize the learned haptic interaction even with variations in body sizes and shapes,
    \item an extension of the BIP framework to enable efficient modeling and inference of high-dimensional, sparse contact forces, and
    \item experimental validation of generalization to previously unseen human interaction partners with varying physical characteristics in a hugging scenario.
\end{itemize}

\begin{figure*}[tb]
	\centering
	\includegraphics[width=0.99\linewidth]{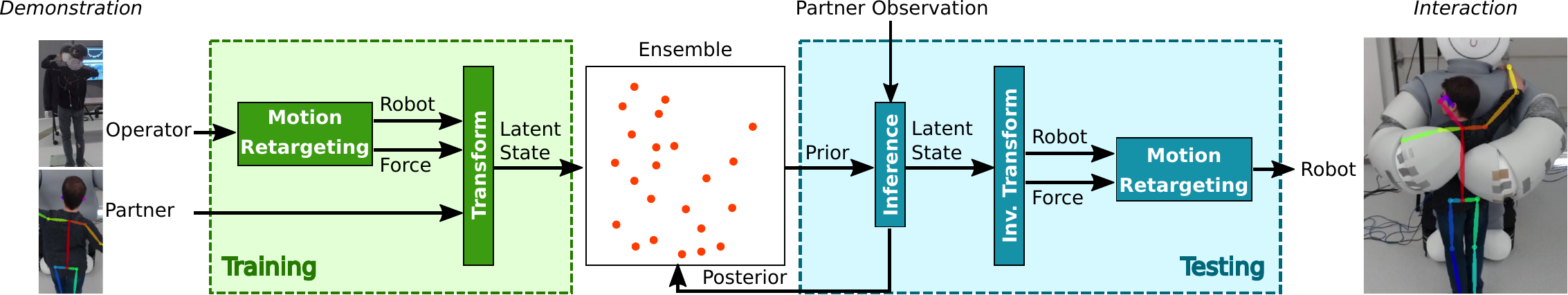}
	\caption{Framework overview. Training demonstrations are used to populate the initial ensemble, which is then updated recursively in testing.}
	\label{fig:overview}
\end{figure*}

\section{Related Work}

\subsection{Social and Physical Human-Robot Interaction}

Intimate, social pHRI such as hugging has been found to have positive effects on the human emotional state~\cite{wada2004effects,rus2015design,shiomi2017robot}.
However, the robot platforms used in these studies all had limited physical capabilities, making it impossible for the robots to provide the human with reciprocal forces.

Block et al.~\cite{block2019softness}, on the other hand, developed a robot based on the PR-2 that can hug a person with significant force.
Their study suggests that hug duration and strength are important factors to realize comfortable robotic hugs.
The caveat is that the robot largely acted independently of the human during experiments.
While the release timing of the hug was adapted in response to user contact, the hug itself was initiated by the robot and the strength was predetermined from three discrete levels.

\subsection{Learning Haptic pHRI}

Recently, robots with whole-body tactile sensing capabilities have been developed for pHRI, in particular, collision response, human-robot communication, and robot behavior development~\cite{argall2010survey}.
Although social pHRI can be considered a form of human-robot communication, most work thus far uses a snapshot~\cite{miyashita2007haptic} or statistic~\cite{noda2012super,kaboli2015humanoids} of tactile information for classifying human intention or behavior.
In contrast, time-series tactile information has largely been limited to non-HRI tasks such as whole-body grasping~\cite{mittendorfer2015realizing} and locomotion~\cite{guadarrama2018enhancing}, although recent studies aim to change this~\cite{bagewadi2019multimodal}.
Advances in sensing may also further this area, as Kim et al.~\cite{kim20153d} proposed to cover a robot with large, airtight, pressure-sensing cavities (in contrast to densely placed small tactile sensors), although this has yet to be applied to pHRI.

Using kinesthetic teaching or traditional haptic devices, one can apply LfD to HRI tasks involving haptic interactions at the end effectors or through an object.
Calinon et al.~\cite{calinon2009learning} presented an LfD framework for teaching a collaborative lifting task by operating a humanoid robot and feeling the interaction force through a PHANToM device.
Peternel et al.~\cite{peternel2013learning} developed a dedicated haptic interface to demonstrate compliant responses to human push and pull of a standing humanoid robot.

\section{Methodology}
\label{sec:method}

In this work, we explicitly model the joint spatiotemporal relationship of the poses and forces between the robot and the partner with Bayesian Interaction Primitives (BIP)~\cite{campbell2017bayesian}.
This relationship is learned from a set of training demonstrations and acts as a source of prior knowledge for the interaction, as depicted in the \emph{Training} block of Fig.~\ref{fig:overview}.

At run-time, the robot is presented with observations of a human partner and, utilizing both the prior knowledge and the current observations, infers (a) the next partner actions, and (b) the appropriate robot response.
This is shown in the \emph{Testing} block of Fig.~\ref{fig:overview}.
The spatiotemporal relationship we model allows us to infer both the robot's poses and forces from those of the partner, allowing responsive behavior to not only discernable movements but also indiscernable ones, such as the strength of the hug.

\subsection{Preliminaries: Bayesian Interaction Primitives~\cite{campbell2017bayesian,campbell2019probabilistic}}
\label{sec:prelim_bip}

We define an interaction $\boldsymbol{Y}$ as a time series of $D$-dimensional sensor observations over time, $\boldsymbol{Y}_{1:T} = [\boldsymbol{y}_1, \dots, \boldsymbol{y}_T] \in \mathbb{R}^{D \times T}$.
The $D$ dimensions include both \emph{observed} degrees of freedom (DoFs) from the human partner and \emph{controlled} DoFs from the the robot.

\subsubsection{Basis Function Decomposition}
\label{sec:prelim_basis_decomp}

In order to decouple the size of the state space from the number of observations, we transform the interaction $\boldsymbol{Y}_{1:T}$ into a time-invariant latent space via basis function decomposition.
Each dimension $d \in D$ of $\boldsymbol{Y}$ is approximated with a weighted linear combination of time-dependent basis functions: $[y^d_{1}, \dots, y^d_{T}] = [\Phi_{\phi(1)}^{d} \boldsymbol{w}^d + \epsilon_y, \dots, \Phi_{\phi(T)}^{d} \boldsymbol{w}^d + \epsilon_y]$, where $\Phi_{\phi(t)}^d \in \mathbb{R}^{1\times B^d}$ is a row vector of $B^d$ basis functions, $\boldsymbol{w}^d \in \mathbb{R}^{B^d \times 1}$, and $\epsilon_y$ is i.i.d. Gaussian noise.
The full latent model is composed of the aggregated weights from each dimension, $\boldsymbol{w} = [\boldsymbol{w}^{1\intercal}, \dots, \boldsymbol{w}^{D\intercal}] \in \mathbb{R}^{1 \times B}$ where $B = \sum_{d}^{D} B^d$,
with the transformation from latent to observation space denoted as $\boldsymbol{y}_t = h(\phi(t), \boldsymbol{w})$.
We define $\phi(t)$ as a linearly interpolated relative phase value with domain $[0, T]$ and codomain $[0, 1]$.

\subsubsection{Spatiotemporal Bayesian Inference}
\label{sec:prelim_filtering}

Informally, BIP seeks to infer the underlying latent model $\boldsymbol{w}$ of an interaction while considering a prior model $\boldsymbol{w}_0$ and a partial sequence of observations, $\boldsymbol{Y}_{1:t}$, such that $\phi(t) \leq 1$ and $T$ is unknown.
Given that $T$ is not known, we must infer both the phase associated with the observations, $\phi(t)$, as well as the phase velocity $\dot{\phi}(t)$ -- how quickly the interaction is progressing.
Our estimate of the underlying latent model contains correlated uncertainties between the individual weights, due to a shared error in the phase estimate.
Intuitively, this is because a temporal error induces a correlated error in spatial terms due to the phase dependency of the basis functions~\cite{campbell2019learning}.
Probabilistically, we represent this with the augmented state vector $\boldsymbol{s} = [\phi, \dot{\phi}, \boldsymbol{w}]$ and the following definition:
\begin{equation}
\label{eq:bip_general}
p(\boldsymbol{s}_t | \boldsymbol{Y}_{1:t}, \boldsymbol{s}_{0}) \propto p(\boldsymbol{y}_{t} | \boldsymbol{s}_t) p(\boldsymbol{s}_t | \boldsymbol{Y}_{1:t-1}, \boldsymbol{s}_{0}).
\end{equation}

Following the ensemble variant of BIP, ensemble Bayesian Interaction Primitives~\cite{campbell2019probabilistic}, the posterior density in Eq.~(\ref{eq:bip_general}) is approximated with a Monte Carlo ensemble which is updated as a two-step recursive filter~\cite{evensen2003ensemble}: the prediction step to propagate each sample forward in time according to a constant velocity state transition function $g(\cdot)$ with process noise $\boldsymbol{Q}$:
\begin{align}
\boldsymbol{x}^j_{t|t-1} &=
g(\boldsymbol{x}^j_{t-1|t-1})
+
\mathcal{N}
\left(0, \boldsymbol{Q}\right),
\label{eq:state_prediction}
\end{align}
and the update step to update the samples given a perturbed measurement $\boldsymbol{\tilde{y}}_t$ and gain coefficient $\boldsymbol{K}$:
\begin{align}
\boldsymbol{x}^j_{t|t} &= \boldsymbol{x}^j_{t|t-1} + \boldsymbol{K} (\boldsymbol{\tilde{y}}_{t} - h(\boldsymbol{x}^j_{t|t-1})).
\label{eq:measurement_update}
\end{align}
The initial ensemble members are obtained directly from the latent space representations of $E$ demonstrations, such that the ensemble consists of $E$ members and $1 \leq j \leq E$:
\begin{equation}
\boldsymbol{x}^j_0 = \left[0, 1/T_j, \boldsymbol{w}_j\right],
\end{equation}
where $\boldsymbol{w}_j$ and $T_j$ are the latent model and length of the $j$-th demonstration respectively.
After each update step, the inferred joint positions and contact forces are computed by transforming the sample mean of the ensemble to the measurement space with
\begin{equation}
\hat{\boldsymbol{y}}_t = h\left( \frac{1}{E} \sum_{j=1}^E 
\boldsymbol{x}^j_{t|t-1} \right)
\end{equation}
and sent to the retargeting controller as reference signals.

\subsection{Feature Selection for Sparse Contact Forces}

The original BIP framework has been successfully applied to pHRI tasks involving haptic information such as pressure at the human soles~\cite{campbell2019probabilistic}.
However, this information has relatively small dimensionality and contains dense pressure data throughout the interaction.
In contrast, the robot employed in this work is equipped with $61$ force sensors over its whole body as discussed in Sec.~\ref{sec:exp_setup} and it is expected that not all of them will be in contact with the partner during an interaction.

One difficulty in working with full-body haptic modalities is that the dimension of the state space tends to increase dramatically.
A large state space is undesirable for several reasons~\cite{campbell2019probabilistic}, most significantly that it can a) increase the error in higher-order statistics that are not tracked
and b) increase the number of ensemble members required to track the true distribution, which negatively impacts computational performance and increases the minimum number of training demonstrations that must be provided.

However, we can exploit the fact that haptic modalities are sparse, both temporally and spatially, in order to reduce the state dimension.
In the case of temporal sparsity, the force sensors will only relay useful information at the time of contact; the time periods before and after are not informative and do not need to be approximated.
In terms of spatial sparsity, some sensors may not experience contact at all depending on the specific interaction.
For example, a hugging motion where both of the partner's arms wrap under the robot's arms will register different contact forces than if the arms wrap over.
Furthermore, this can vary between person to person as physical characteristics such as height influence contact location and strength.

To utilize temporal sparsity, we employ a non-uniform distribution of basis functions across a subregion of the phase domain, unlike in prior works~\cite{campbell2017bayesian, campbell2019probabilistic} where a uniform distribution over the full domain of $[0, 1]$ was utilized.
The basis space is found via Orthogonal Least Squares (OLS)~\cite{chen1991orthogonal}, as our basis decomposition shares many similarities with a $1$-layer Radial Basis Function network.
This allows us to reduce the number of required basis functions while still adequately covering the informative interaction periods.

Spatial sparsity, however, requires a different approach as we cannot know which sensors will be useful for a given interaction in advance.
We therefore employ a feature selection method based on mutual information~\cite{peng2005feature, vergara2014review} to maximize the dependency between the input and output force sensor DoFs.
Some care must be taken as there are a limited set of training demonstrations available and we are selecting based on continuous state variables (basis function coefficients), however, this can be overcome through binning~\cite{kraskov2004estimating}, K-nearest neighbors~\cite{ross2014mutual}, or Parzen windows~\cite{kwak2002input}.
In our case, we employ mutual information estimation based on binning and incrementally select features until there is no significant improvement in mutual information, based on a desired threshold ($> 0.07$).
We opt to use mutual information rather than other standard measures as some of the force sensors experience false positives due to deformation caused by the robot's own movements.

By considering the physical layout of the robot, we can further reduce the state dimension.
In general, tactile sensors have a limited sensing range and are used in large groups to provide greater surface coverage.
However, when a human exerts a contact force to this region we do not care whether it activated the first or the tenth sensor of a group, we simply care that force was exerted on this region.
This allows us to reduce an entire group of force sensors to a single DoF by simply taking the maximum force value of all the sensors in that group at any given point in time.

\section{Experimental Setup}
\label{sec:exp_setup}

In this work, we collect human-robot demonstrations with a teleoperated bimanual robot as it is difficult to demonstrate intimate interactions through direct kinesthetic teaching.
Compared to obtaining data from human-human demonstrations, this method has the advantage of directly modeling the motion and force which the robot is physically capable of as well as genuine human reactions.

\begin{figure}[tb]
    \centering
    \includegraphics[width=0.6\linewidth]{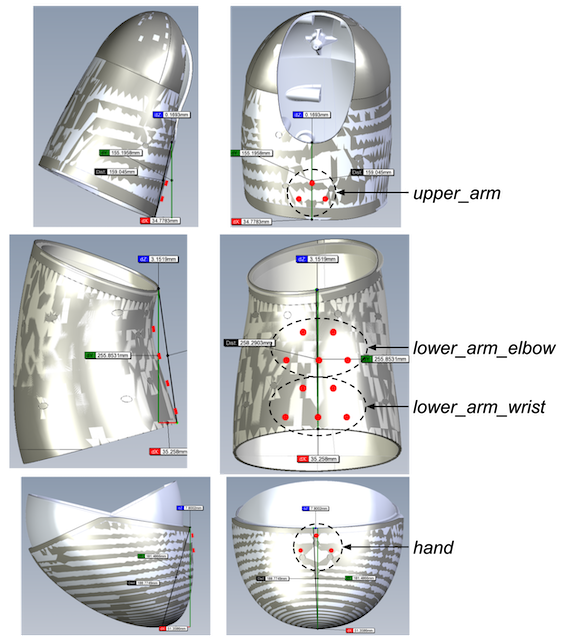}
    \caption{Force sensor placement on the robot arm. The dotted circles denote the grouping corresponding to the force sensors attached to the operator.}
    \label{fig:arm_sensor_placements}
\end{figure}
\begin{figure}[tb]
    \centering
    \includegraphics[width=0.49\linewidth]{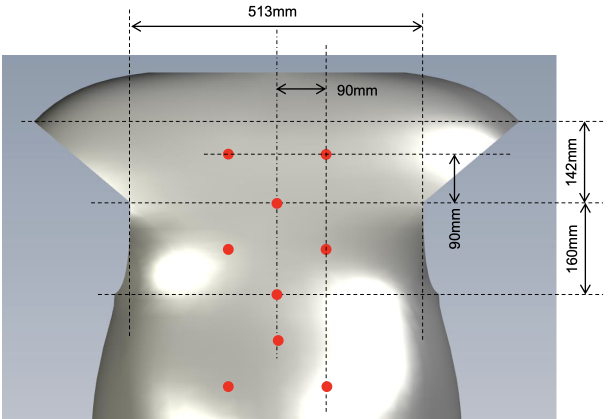}
    \includegraphics[width=0.49\linewidth]{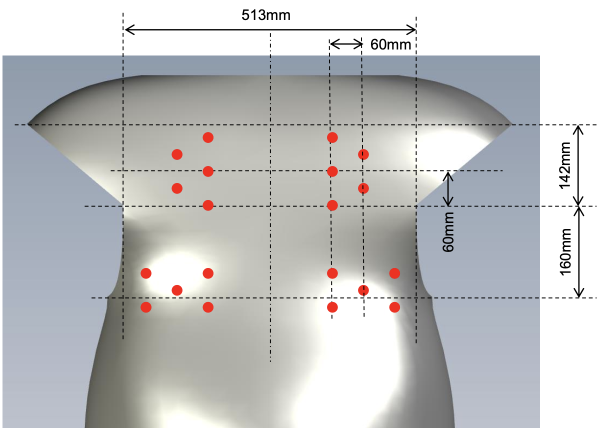}
    \caption{Force sensor placement on the robot's chest (left) and back (right).}
    \label{fig:torso_sensor_placements}
\end{figure}

\subsection{Robot Hardware and Teleoperation System~\cite{kaplish2019motion}}

We use a teleoperated bimanual robot consisting of two torque-controlled Franka Emika 7-DOF arms~\cite{bib-franka-emika} wrapped in a soft, padded exterior.
A set of $61$ force sensors~\cite{bib-touchence} are embedded into the padded exterior of the robot, providing tactile information on the robot's arms (Fig.~\ref{fig:arm_sensor_placements}), chest (Fig.~\ref{fig:torso_sensor_placements} left), and back (Fig.~\ref{fig:torso_sensor_placements} right).

During teleoperation, the operator hugs a static mannequin while wearing a suit equipped with 8 IMUs~\cite{bib-imu} (2 on the back and 1 on each hand, forearm, and upper arm) as well as 8 force sensors (1 on each hand, 2 on each forearm, and 1 on each upper arm).
Because the robot's arms are much larger than the human's, each force sensor on the operator corresponds to a group of 3--5 sensors placed on a topologically similar location on the robot arm, as shown in Fig.~\ref{fig:arm_sensor_placements}.
This placement ensures that contact forces are detected even with variation in motion and body shape.
We use the raw output of the force sensors (via an 8-bit A/D converter) directly as a reference signal, possible only because the human and robot use the same type of sensor.

A motion retargeting controller has been developed~\cite{kaplish2019motion} to adapt the operator motion such that the contact timing, states, and magnitudes on the robot approximately match those of the operator, thereby accounting for variations in body sizes and shapes.
This retargeting controller is used in both training, in which the operator's force on the mannequin is matched, and in testing, in which the inferred contact force generated by BIP is matched.

\begin{figure}[tb]
    \centering
    \includegraphics[width=1\linewidth]{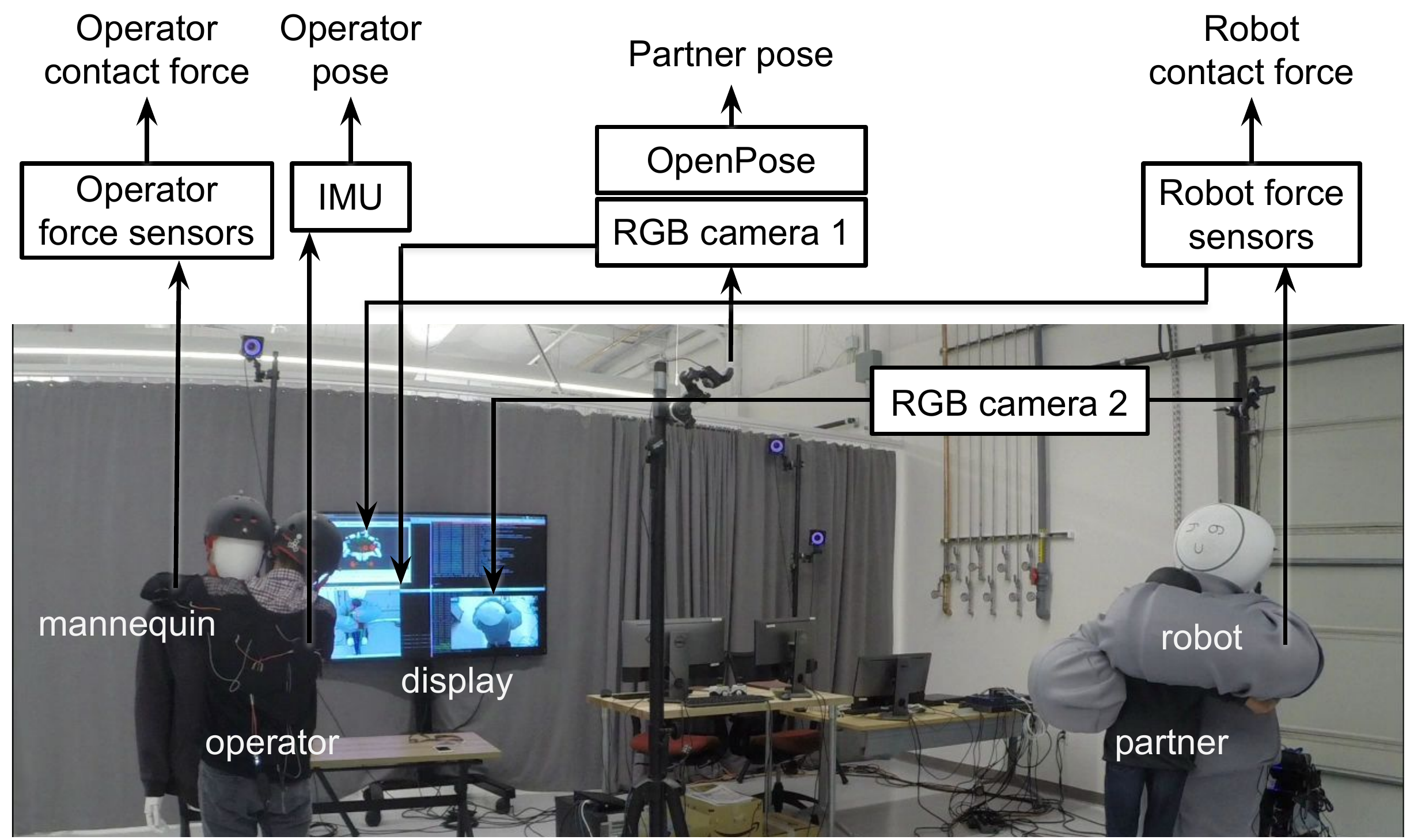}
    \caption{Demonstration data collection through teleoperation.}
    \label{fig:teleop_demonstration_setup}
\end{figure}

\begin{figure*}[tb]
    \centering
    \includegraphics[width=0.19\linewidth]{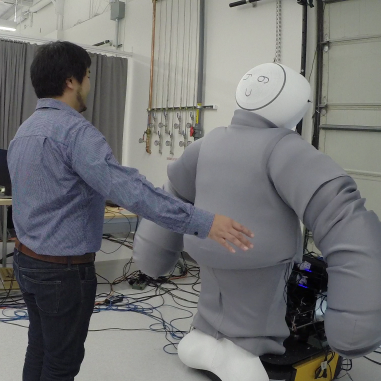}
    \includegraphics[width=0.19\linewidth]{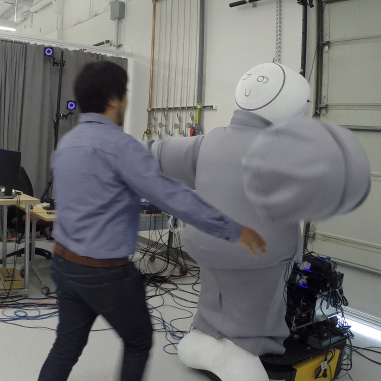}
    \includegraphics[width=0.19\linewidth]{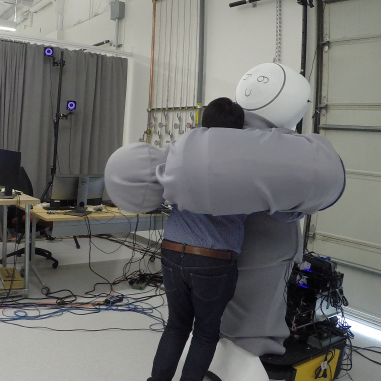}
    \includegraphics[width=0.19\linewidth]{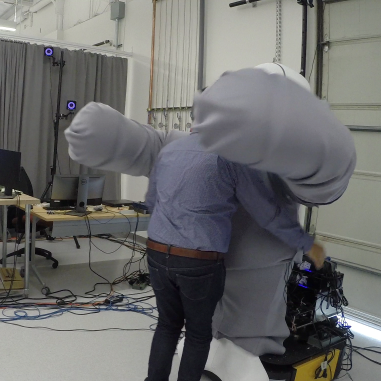}
    \includegraphics[width=0.19\linewidth]{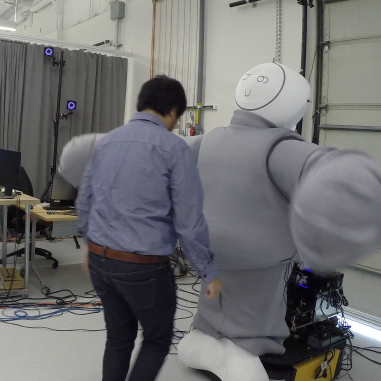}
    \caption{Snapshots from a hug interaction.}
    \label{fig:interaction_example}
\end{figure*}

\begin{figure}[tb]
    \centering
    \includegraphics[width=0.9\linewidth, trim={0cm 1cm 0cm 1.5cm},clip]{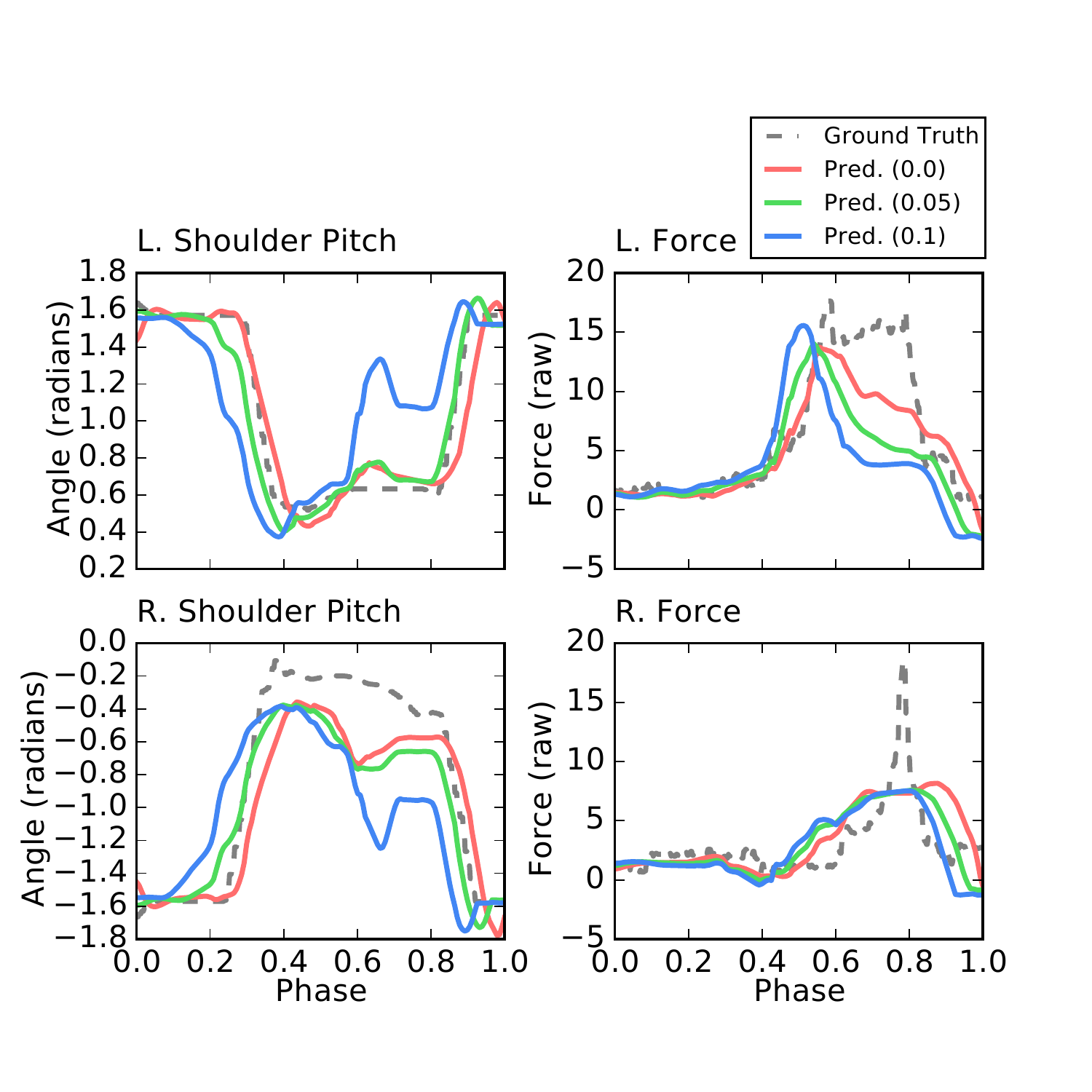}
    \caption{Left: shoulder pitch joint angle in the robot's left (top) and right (bottom) arms. The dashed gray line indicates the ground truth, while the solid red line indicates the prediction with a phase look-ahead of $0.0$, green with $0.05$, and blue with $0.1$. Right: the force ground truth and predicted values for the mean wrist force sensors in the left and right arms.}
    \label{fig:dynamic_results1}
\end{figure}

\begin{figure}[tb]
    \centering
    \includegraphics[width=0.95\linewidth, trim={2cm 0 2cm 0.5cm},clip]{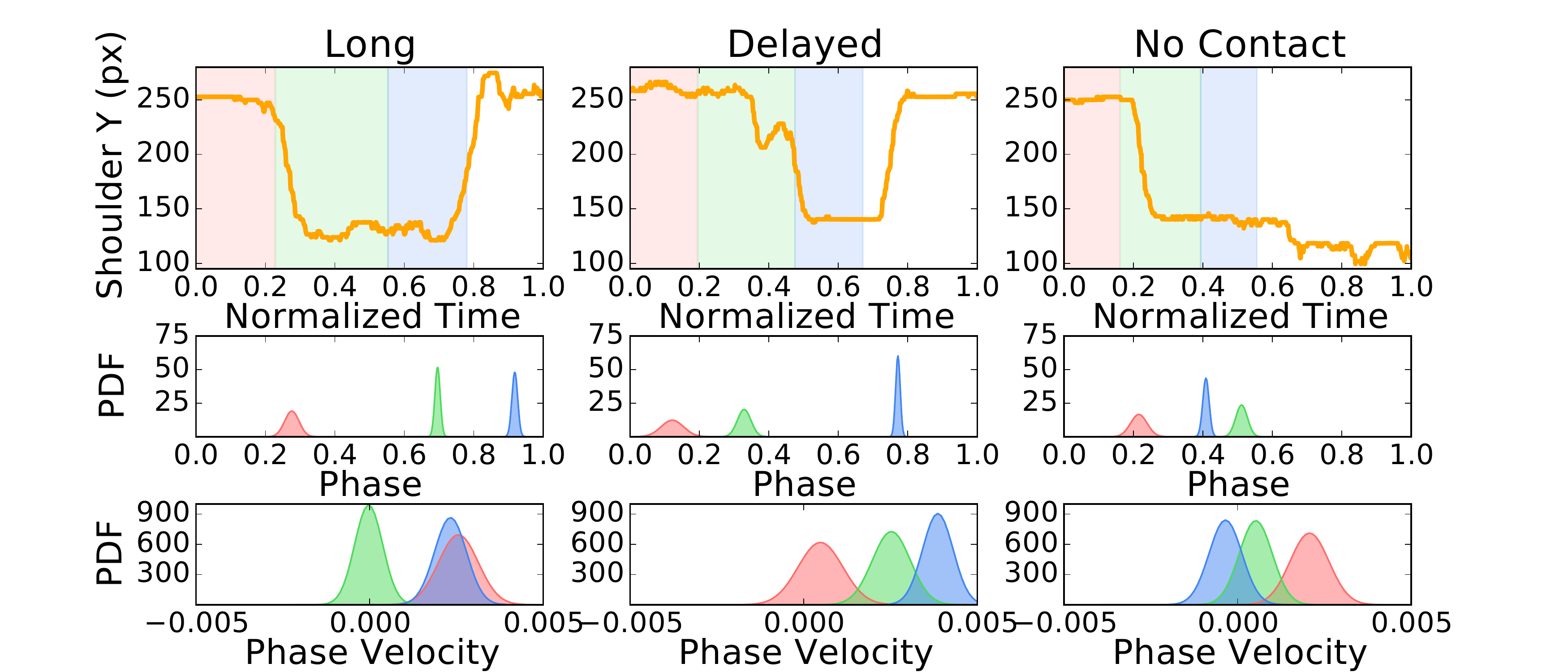}
    \caption{Top: the position of the human's right shoulder along the $y$-axis as determined by pose tracking. The red shaded region indicates the portion of the interaction before movement has begun, the green region the portion after contact has been made, and the blue region the portion where the human is withdrawing from the hug. Middle: the phase distribution corresponding to the end of each region. Bottom: the phase velocity distribution.}
    \label{fig:dynamic_results2}
\end{figure}

\begin{figure}[tb]
    \centering
    \includegraphics[width=0.32\linewidth]{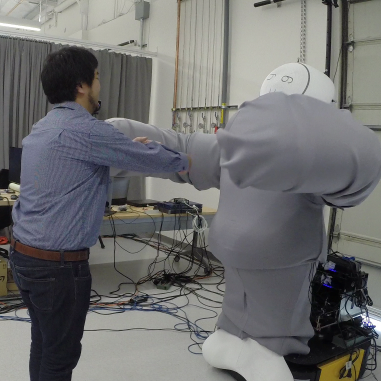}
    \includegraphics[width=0.32\linewidth]{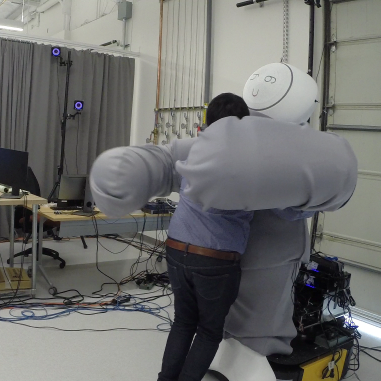}
    \includegraphics[width=0.32\linewidth]{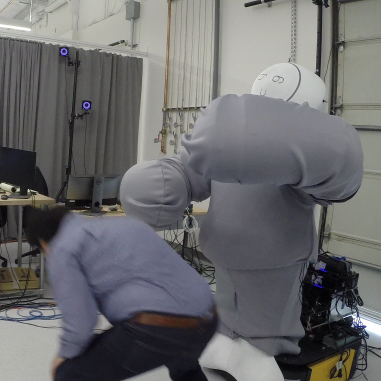}
    \caption{Edge cases. 
    Left: hugging the air; the hug did not complete because the phase did not proceed further.
    Center: delay before hugging; hug was successful because the model correctly recognized the beginning of a hug. 
    Right: hugging without making contact; hug failed because the model received conflicting information.}
    \label{fig:edge_cases}
\end{figure}

\begin{figure}[tb]
    \centering
    \includegraphics[width=0.90\linewidth, trim={0cm 0.5cm 0cm 1.25cm},clip]{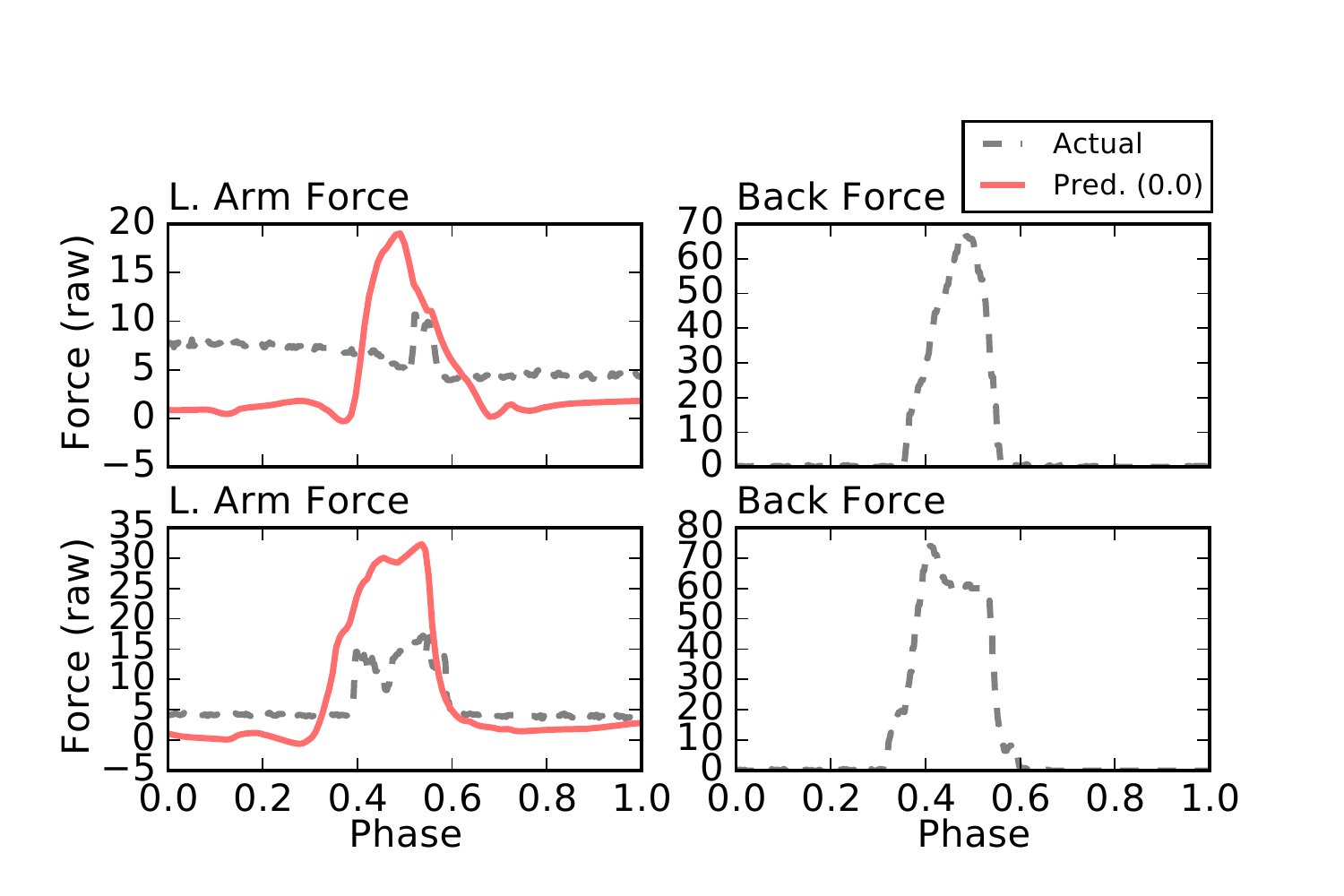}
    \caption{Top: the contact forces for the mean of the wrist force sensors in the left (left) and a back sensor (right) for a single interaction. The dashed gray line indicates the actual force and the solid red line the predicted. Bottom: the contact forces for a different interaction.}
    \label{fig:dynamic_results3}
\end{figure}

\subsection{Demonstration Data Collection}
\label{sec:exp_setup_hw_teleop}

Figure~\ref{fig:teleop_demonstration_setup} shows the setup for collecting learning data using the teleoperated robot.
In addition to the sensors attached to the operator and robot, we use two RGB cameras.
The first is mounted at a fixed point above and behind the partner and used to estimate the partner's 2D pose at $30$Hz using the OpenPose library~\cite{cao2018openpose}.
To help the operator decide the motion and timing, the second camera is fixed behind the robot and allows the operator to view the interaction with the partner.
The pose estimate of the partner is visualized for the operator's benefit, and the force sensor readings are overlaid as spheres on a model of the robot.
Their size and color are dictated by the magnitude of the current reading, such that the operator can compensate for the lack of haptic feedback indicating the force exerted by the partner.

In summary, we collect the following data for training:
\begin{itemize}
\item Operator: motion (8 IMUs) and forces (8 sensors),
\item Robot: 12 joint position commands generated by motion retargeting and contact forces (61 sensors), and
\item Partner: pose (RGB camera and OpenPose).
\end{itemize}
Although we also measure the positions of the robot, partner, operator, and mannequin using an optical motion capture system, we ultimately did not use this data as the partner's distance from the robot was indirectly inferred based on the partner's vertical location in the image.

\subsection{Training and Testing}

We recruited $6$ total participants to hug the robot, $4$ of whom participated in training and all $6$ in testing (a combination of offline and online); this allows us to test generalization capabilities as $2$ of the testing participants were not previously seen.
During training, the participants stood approximately $2$m away from the robot and were instructed to step forward and initiate a hug by raising their arms.
The robot, controlled by another person via teleoperation as described in Sec.~\ref{sec:exp_setup_hw_teleop}, responded by raising its arms so that the hug may proceed.
The participants were also instructed to take the lead in disengaging from the hug.
That is, when the participant ceased to apply pressure to the robot, indicating that the hug was over, the operator lowered the robot's arms such that the participant could step back and conclude the hug.
Furthermore, the participants were given instructions to attempt to match the strength of the robot's hug.
This instruction was left intentionally vague such that each participant would naturally apply an amount of force appropriate to their own preferences in response to the force exerted by the robot.
This was repeated $25$ times for each participant resulting in a total of $150$ training demonstrations, although we only used $121$ of these as the remainder did not experience significant contact forces.
The same operator controlled the robot for all participants and the BIP prior was trained using all $121$ training demonstrations.
The online tests were conducted with a model that utilized both force sensor grouping (spatial sparsity) and a non-uniform basis space selected with OLS (temporal sparsity).

During online testing, the participants were instructed to initiate and conclude the hug, as in training, except now with no teleoperation and no instruction to match the force of the robot.
Each participant performed six long hugs, six short hugs, and two hugs each for the following edge cases: doing nothing, delaying before hugging, delaying after raising arms, hugging the air without moving, and hugging the robot without making contact.
Online testing was performed with $3$ participants -- $2$ of whom were novel and did not participate in training, and $1$ who had participated in training -- for a total of $66$ hugs.

In addition to online testing we also conducted offline testing in which the BIP model's inference accuracy was compared against the ground truth data from demonstrations of all $4$ training participants.
This analysis was conducted using 10-fold cross validation such that the BIP was trained using $108$ demonstrations at a time.
Four model variations were tested: one which used all input force DoFs (All), one with input force DoFs selected through mutual information feature selection (MIFS), one with force sensor grouping without MIFS (Group), and one with both force sensor grouping and non-uniform basis space (Group + OLS).

\section{Results and Discussion}

\begin{table}[]
    \vspace*{2mm}
    \caption{The mean absolute error (MAE) values for the predicted joint (radians), left arm force (raw), and right arm force (raw) values for a phase look-ahead of $0.0$, $0.05$, and $0.1$ computed using 10-fold cross validation. A green box represents the best method and a red box any method which \emph{is} statistically worse than the best method (Mann-Whitney U, $p < 0.05$).}
    \label{tab:results_offline_mae}
    \centering
        \begin{tabular}{cccccc}
             & & \thead{All} & \thead{MIFS} & \thead{Group} & \thead{Group + OLS}  \\
             \cmidrule(lr){2-6}\morecmidrules\cmidrule(lr){2-6}
             & Dimension & 664 & 446 & 392 & 378 \\
             \cmidrule(lr){2-6}
             \parbox[t]{2mm}{\multirow{3}{*}{\rotatebox[origin=c]{90}{0.00}}} & Joints & 0.191 & \cellcolor{green!15} 0.180 & 0.183 & 0.180 \\
             & Left & \cellcolor{green!15} 7.224 & 7.580 & 8.011 & 7.937 \\
             & Right & \cellcolor{green!15} 3.760 & 4.163 & 3.936 & 3.868 \\
             \cmidrule(lr){2-6}
             \parbox[t]{2mm}{\multirow{3}{*}{\rotatebox[origin=c]{90}{0.05}}} & Joints & \cellcolor{red!15} 0.213 & \cellcolor{green!15} 0.201 & 0.209 & 0.205 \\
             & Left & 7.992 & \cellcolor{green!15} 7.965 & 8.257 & 8.077 \\
             & Right & 4.058 & 4.110 & 3.897 & \cellcolor{green!15} 3.850 \\
             \cmidrule(lr){2-6}
             \parbox[t]{2mm}{\multirow{3}{*}{\rotatebox[origin=c]{90}{0.10}}} & Joints & 0.250 & \cellcolor{green!15} 0.238 & 0.247 & 0.243 \\
             & Left & 9.03 & 8.378 & 8.257 & \cellcolor{green!15} 8.180 \\
             & Right & 4.355 & 4.061 & 3.921 & \cellcolor{green!15} 3.867 \\
             \cmidrule(lr){2-6}\morecmidrules\cmidrule(lr){2-6}
        \end{tabular}
\end{table}

Figure~\ref{fig:interaction_example} shows snapshots from a successful hug. 
The supplemental video shows select demonstrations and online interactions including the edge cases.
During online testing, we achieved an approximately $82\%$ ($54/66$) qualitative success rate which we define as the robot hugging the human participant and responding to their cues; this is by definition a subjective metric but it is a useful frame of reference.
Breaking this down further, novel participants were hugged successfully ${\sim}80\%$ (35/44) of the time while the sole participant who also trained was successful ${\sim}86\%$ (19/22) of the time.

Table~\ref{tab:results_offline_mae} shows the mean absolute error results for the inferred joint positions and contact forces in the left and right arms in offline testing.
We show that despite our proposed method (Group + OLS) having the smallest state dimension, it never performs significantly worse than any other method.
This indicates that we can leverage the sparsity of contact forces in order to reduce the state dimension without negatively impacting inference accuracy.
We note that the smaller MAE values for the right arm forces do not indicate a better prediction, simply that they experienced less pressure; we emphasize the comparison between methods and not output DoFs.

We also introduce the notion of a phase look-ahead value in these results, which is a non-negative offset applied to the inferred phase.
A phase look-ahead of $0.0$ means we predict values for the currently estimated phase; a look-ahead of $0.1$ means we are predicting values $10\%$ (relative to phase) into the future.
This look-ahead is vital for natural interactions, as it allows us to overcome phase lag and produce more responsive robot actions.
However, as Table~\ref{tab:results_offline_mae} demonstrates, a higher look-ahead value results in higher inference error.
This is further demonstrated in Fig.~\ref{fig:dynamic_results1}.
While all look-ahead values result in accurate responses, higher look-ahead values result in temporally earlier responses at the cost of larger errors (especially noticeable with $0.1$ look-ahead error).

Figure~\ref{fig:dynamic_results2} shows the phase estimate distributions for various edge cases conducted during online testing.
The BIP model used in this work is robust to temporal variance which is exhibited in the phase and phase velocity adaptation here.
Most significantly, we show that when there is a long hug duration the phase velocity drops to $0$ and the interaction essentially halts (left column, green), until the human partner withdraws their arms (left column, blue).
Similarly, a delay at the beginning of the interaction results in a phase velocity of $0$ (middle, red), ceasing the temporal progression of the interaction until the human begins to move (middle, green).
In the case of no contact, the interaction progresses until the point where contact pressure is expected (right, green).
As this pressure never comes, the interaction halts indefinitely until cancelled (right, blue).
Snapshots from some edge cases are shown in Fig.~\ref{fig:edge_cases}.

Figure~\ref{fig:dynamic_results3} demonstrates a weakness in the BIP interaction model: it does not account for external physical limitations.
In this case, two different interactions are shown in which the inferred contact force for the left arm is driven by a force on the back, however, the actual resulting left contact force is unable to realize the inferred value.
This can be due to multiple reasons, including joint angle limitations, limited force sensor coverage, collision between hands, and unexpected movement from the human.

\section{Conclusions}

In this paper, we presented an LfD framework for teaching intimate, social pHRI.
Intimate interactions involve direct, wide-area contacts between the robot and human partner that are difficult to model and control with traditional pHRI frameworks that focus on interactions at the end effectors.
In social interactions, this ability to sense and react to subtle whole-body tactile information is essential.

Our framework solves these issues by incorporating whole-body haptic information into the model, and by building a model that captures the relationship between the haptic information and robot motion.
We extend the existing BIP framework to allow high-dimensional, sparse tactile data seen in intimate physical interactions, where contact forces are relevant only in a part of the interaction and the set of active sensors may be different among demonstrations.
We also combine the extended BIP algorithm with a motion retargeting controller to realize the predicted motion and force across different human body shapes and sizes.

We experimentally validated the accuracy and generalizability of our framework using hugging as a case study.
We trained the BIP model with 121 demonstrations collected from 4 participants interacting with a teleoperated bimanual robot with 61 force sensors, and tested the model offline with all 4 training participants and online with 1 training participant as well as 2 novel partners.
The model was able to generate reasonable robot responses for every standard hug performed by all participants, as well as edge cases such as delaying before hugging (in online testing).

Informal comments from the participants suggested that the robot response with a small phase look-ahead appeared more natural and therefore easier to interact with, possibly due to slow response of the motion retargeting controller.
Future work includes a perception study to identify the optimal values of various parameters, including the phase look-ahead, that realize the most emotionally effective interactions.

\bibliographystyle{IEEEtran}
\bibliography{references}

\begin{thebibliography}{10}
\providecommand{\url}[1]{#1}
\csname url@rmstyle\endcsname
\providecommand{\newblock}{\relax}
\providecommand{\bibinfo}[2]{#2}
\providecommand\BIBentrySTDinterwordspacing{\spaceskip=0pt\relax}
\providecommand\BIBentryALTinterwordstretchfactor{4}
\providecommand\BIBentryALTinterwordspacing{\spaceskip=\fontdimen2\font plus
\BIBentryALTinterwordstretchfactor\fontdimen3\font minus
  \fontdimen4\font\relax}
\providecommand\BIBforeignlanguage[2]{{%
\expandafter\ifx\csname l@#1\endcsname\relax
\typeout{** WARNING: IEEEtran.bst: No hyphenation pattern has been}%
\typeout{** loaded for the language `#1'. Using the pattern for}%
\typeout{** the default language instead.}%
\else
\language=\csname l@#1\endcsname
\fi
#2}}

\bibitem{argall2010survey}
B.~D. Argall and A.~G. Billard, ``A survey of tactile human--robot
  interactions,'' \emph{Robotics and autonomous systems}, vol.~58, no.~10, pp.
  1159--1176, 2010.

\bibitem{campbell2017bayesian}
J.~Campbell and H.~Ben~Amor, ``Bayesian interaction primitives: A slam approach
  to human-robot interaction,'' in \emph{Conference on Robot Learning}, 2017,
  pp. 379--387.

\bibitem{campbell2019probabilistic}
J.~Campbell, S.~Stepputtis, and H.~Ben~Amor, ``Probabilistic multimodal
  modeling for human-robot interaction tasks,'' in \emph{Robotics: Science and
  Systems}, 2019.

\bibitem{campbell2019learning}
J.~Campbell, A.~Hitzmann, S.~Stepputtis, S.~Ikemoto, K.~Hosoda, and
  H.~Ben~Amor, ``Learning interactive behaviors for musculoskeletal robots
  using bayesian interaction primitives,'' in \emph{IEEE/RSJ International
  Conference on Intelligent Robots and Systems}, 2019.

\bibitem{kaplish2019motion}
A.~Kaplish and K.~Yamane, ``Motion regargeting and control for teleoperated
  physical human-robot interaction,'' in \emph{IEEE-RAS International
  Conference on Humanoid Robots}, 2019 (in press).

\bibitem{wada2004effects}
K.~Wada, T.~Shibata, T.~Saito, and K.~Tanie, ``Effects of robot-assisted
  activity for elderly people and nurses at a day service center,''
  \emph{Proceedings of the IEEE}, vol.~92, no.~11, pp. 1780--1788, 2004.

\bibitem{rus2015design}
D.~Rus and M.~T. Tolley, ``Design, fabrication and control of soft robots,''
  \emph{Nature}, vol. 521, no. 7553, p. 467, 2015.

\bibitem{shiomi2017robot}
M.~Shiomi, A.~Nakata, M.~Kanbara, and N.~Hagita, ``A robot that encourages
  self-disclosure by hug,'' in \emph{International Conference on Social
  Robotics}.\hskip 1em plus 0.5em minus 0.4em\relax Springer, 2017, pp.
  324--333.

\bibitem{block2019softness}
A.~E. Block and K.~J. Kuchenbecker, ``Softness, warmth, and responsiveness
  improve robot hugs,'' \emph{International Journal of Social Robotics},
  vol.~11, no.~1, pp. 49--64, 2019.

\bibitem{miyashita2007haptic}
T.~Miyashita, T.~Tajika, H.~Ishiguro, K.~Kogure, and N.~Hagita, ``Haptic
  communication between humans and robots,'' in \emph{Robotics Research}.\hskip
  1em plus 0.5em minus 0.4em\relax Springer, 2007, pp. 525--536.

\bibitem{noda2012super}
T.~Noda, T.~Miyashita, H.~Ishiguro, and N.~Hagita, ``Super-flexible skin
  sensors embedded on the whole body, self-organizing based on haptic
  interactions,'' \emph{Human-Robot Interaction in Social Robotics}, p. 183,
  2012.

\bibitem{kaboli2015humanoids}
M.~Kaboli, A.~Long, and G.~Cheng, ``Humanoids learn touch modalities
  identification via multi-modal robotic skin and robust tactile descriptors,''
  \emph{Advanced Robotics}, vol.~29, no.~21, pp. 1411--1425, 2015.

\bibitem{mittendorfer2015realizing}
P.~Mittendorfer, E.~Yoshida, and G.~Cheng, ``Realizing whole-body tactile
  interactions with a self-organizing, multi-modal artificial skin on a
  humanoid robot,'' \emph{Advanced Robotics}, vol.~29, no.~1, pp. 51--67, 2015.

\bibitem{guadarrama2018enhancing}
J.~R. Guadarrama-Olvera, F.~Bergner, E.~Dean, and G.~Cheng, ``Enhancing biped
  locomotion on unknown terrain using tactile feedback,'' in \emph{2018
  IEEE-RAS 18th International Conference on Humanoid Robots (Humanoids)}.\hskip
  1em plus 0.5em minus 0.4em\relax IEEE, 2018, pp. 1--9.

\bibitem{bagewadi2019multimodal}
K.~Bagewadi, J.~Campbell, and H.~B. Amor, ``Multimodal dataset of human-robot
  hugging interaction,'' \emph{arXiv preprint arXiv:1909.07471}, 2019.

\bibitem{kim20153d}
J.~Kim, A.~Alspach, and K.~Yamane, ``3d printed soft skin for safe human-robot
  interaction,'' in \emph{2015 IEEE/RSJ International Conference on Intelligent
  Robots and Systems (IROS)}.\hskip 1em plus 0.5em minus 0.4em\relax IEEE,
  2015, pp. 2419--2425.

\bibitem{calinon2009learning}
S.~Calinon, P.~Evrard, E.~Gribovskaya, A.~Billard, and A.~Kheddar, ``Learning
  collaborative manipulation tasks by demonstration using a haptic interface,''
  in \emph{2009 International Conference on Advanced Robotics}.\hskip 1em plus
  0.5em minus 0.4em\relax IEEE, 2009, pp. 1--6.

\bibitem{peternel2013learning}
L.~Peternel and J.~Babi{\v{c}}, ``Learning of compliant human--robot
  interaction using full-body haptic interface,'' \emph{Advanced Robotics},
  vol.~27, no.~13, pp. 1003--1012, 2013.

\bibitem{evensen2003ensemble}
G.~Evensen, ``The ensemble kalman filter: Theoretical formulation and practical
  implementation,'' \emph{Ocean dynamics}, vol.~53, no.~4, pp. 343--367, 2003.

\bibitem{chen1991orthogonal}
S.~Chen, C.~F. Cowan, and P.~M. Grant, ``Orthogonal least squares learning
  algorithm for radial basis function networks,'' \emph{IEEE Transactions on
  neural networks}, vol.~2, no.~2, pp. 302--309, 1991.

\bibitem{peng2005feature}
H.~Peng, F.~Long, and C.~Ding, ``Feature selection based on mutual information:
  criteria of max-dependency, max-relevance, and min-redundancy,'' \emph{IEEE
  Transactions on Pattern Analysis \& Machine Intelligence}, no.~8, pp.
  1226--1238, 2005.

\bibitem{vergara2014review}
J.~R. Vergara and P.~A. Est{\'e}vez, ``A review of feature selection methods
  based on mutual information,'' \emph{Neural computing and applications},
  vol.~24, no.~1, pp. 175--186, 2014.

\bibitem{kraskov2004estimating}
A.~Kraskov, H.~St{\"o}gbauer, and P.~Grassberger, ``Estimating mutual
  information,'' \emph{Physical review E}, vol.~69, no.~6, p. 066138, 2004.

\bibitem{ross2014mutual}
B.~C. Ross, ``Mutual information between discrete and continuous data sets,''
  \emph{PloS one}, vol.~9, no.~2, p. e87357, 2014.

\bibitem{kwak2002input}
N.~Kwak and C.-H. Choi, ``Input feature selection by mutual information based
  on parzen window,'' \emph{IEEE Transactions on Pattern Analysis \& Machine
  Intelligence}, no.~12, pp. 1667--1671, 2002.

\bibitem{bib-franka-emika}
{Franka Emika GmbH}, ``Introducing the {Franka Emika} robot,''
  https://www.franka.de/.

\bibitem{bib-touchence}
{Touchence Inc.}, ``{ShokacPot/ShokacCube} product outline,''
  http://www.touchence.jp/en/cube/index.html.

\bibitem{bib-imu}
InertialLabs, ``Miniature and subminiature orientation sensors,''
  https://inertiallabs.com/3os.html.

\bibitem{cao2018openpose}
Z.~Cao, G.~Hidalgo, T.~Simon, S.-E. Wei, and Y.~Sheikh, ``Open{P}ose: realtime
  multi-person 2{D} pose estimation using {P}art {A}ffinity {F}ields,'' in
  \emph{arXiv preprint arXiv:1812.08008}, 2018.

\end{thebibliography}

\end{document}